\documentclass[conference]{IEEEtran}
\IEEEoverridecommandlockouts

\usepackage{cite}
\usepackage{amsmath,amssymb,amsfonts}
\usepackage{algorithmic}
\usepackage{graphicx}
\usepackage{textcomp}
\usepackage{xcolor}
\usepackage{booktabs}
\usepackage{url}
\usepackage{tikz}

\usetikzlibrary{arrows.meta,positioning,shapes.geometric,fit,backgrounds,calc,shadows}
\usepackage{hyperref}
\hypersetup{
    colorlinks=true,
    linkcolor=blue,
    filecolor=magenta,
    urlcolor=blue,
    citecolor=blue,
}

\def\BibTeX{{\rm B\kern-.05em{\sc i\kern-.025em b}\kern-.08em
    T\kern-.1667em\lower.7ex\hbox{E}\kern-.125emX}}

\definecolor{osfblue}{RGB}{43,98,160}
\definecolor{osfbluepale}{RGB}{223,232,243}
\definecolor{pulseorange}{RGB}{218,123,38}
\definecolor{pulseorangepale}{RGB}{248,233,217}
\definecolor{bridgegreen}{RGB}{52,138,93}
\definecolor{bridgegreenpale}{RGB}{225,238,231}
\definecolor{outpurple}{RGB}{120,80,160}
\definecolor{outpurplepale}{RGB}{236,229,243}
\definecolor{anomred}{RGB}{200,70,70}
\definecolor{regiongray}{RGB}{120,120,120}

\definecolor{mbblue}{RGB}{224,235,247}
\definecolor{mbgreen}{RGB}{232,244,234}
\definecolor{mborange}{RGB}{255,242,224}
\definecolor{mbpurple}{RGB}{244,235,255}
\definecolor{mbgray}{RGB}{242,242,242}

\begin{document}

\title{MagBridge-Battery: A Synthetic Bridge Dataset for Li-ion Magnetometry and State-of-Health Diagnostics}

\author{
\IEEEauthorblockN{Sakthi Prabhu Gunasekar and Prasanna Kumar Rangarajan}
\IEEEauthorblockA{
Dept.\ of Computer Science \& Engineering\\
Amrita School of Computing\\
Amrita Vishwa Vidyapeetham, India\\
ORCID: \href{https://orcid.org/0009-0006-0153-5674}{0009-0006-0153-5674},
\href{https://orcid.org/0000-0001-6103-259X}{0000-0001-6103-259X}
}
}
\maketitle

\begin{abstract}
Battery health diagnostics today rely overwhelmingly on electrochemical signals at the cell terminals. A parallel literature has shown that magnetic sensing can resolve information that terminal-only measurements miss, but the central obstacle to method development in this area is the absence, to the best of our knowledge, of public battery magnetic-measurement datasets paired with degradation labels. We release \textbf{MagBridge-Battery v1.0}, a synthetic dataset of 6{,}760 magnetic-field signatures bridging real magnetic morphology from the Mohammadi--Jerschow Open Science Framework (OSF) archive with real state-of-health (SOH) labels from the PulseBat dataset. The release contains 5{,}600 PulseBat-conditioned grounded samples, 600 synthetic sensor-anomaly samples derived from clean parents, and 560 low-voltage Regime-B extrapolation samples. A cell-disjoint, parent-child-leakage-free primary benchmark split is verified to contain zero overlapping cells, zero cross-split parent-child pairs, and zero sample-ID overlap. We define three primary benchmark tasks (SOH regression, second-life classification, anomaly detection) and one auxiliary anomaly-subtype classification task, and validate the dataset with a controlled ablation suite: a label-shuffle ablation collapses SOH regression from $R^2 \approx 0.77$ to $\approx 0$, confirming that the bridge encodes input SOH non-trivially rather than producing label-aligned artefacts. The dataset is released on Zenodo under CC-BY-4.0; the bridge code and benchmark suite are released under Apache-2.0. This work bridges a gap in public data for magnetic-sensing battery diagnostics while paired magnetic--electrochemical measurements remain scarce.
\end{abstract}

\begin{IEEEkeywords}
battery diagnostics, magnetic sensing, state-of-health estimation, synthetic dataset, benchmark, second-life batteries, out-of-distribution evaluation, lithium iron phosphate, leakage-safe split
\end{IEEEkeywords}

\section{Motivation}
\label{sec:motivation}

Battery health diagnostics today rely overwhelmingly on \emph{electrochemical signals}: voltage, current, temperature, and impedance measurements taken at the cell terminals. The public datasets that drive method development reflect this. NASA, Oxford, CALCE, Stanford/Severson, MATR, HUST, XJTU, and PulseBat all provide rich electrochemical time-series at varying degrees of degradation, with well-characterised SOH labels and, in many cases, full degradation trajectories~\cite{severson2019,tao2024pulsebat}. These datasets have enabled a generation of work on early-life prediction, capacity-fade estimation, and second-life classification.

What none of them capture is the \emph{spatial distribution of current and magnetisation inside the cell}. A terminal-only measurement is, by construction, blind to localised hot-spots of charge storage, inhomogeneous redox at electrode surfaces, dendrite formation, and the kinds of internal defects that emerge during ageing or that signal manufacturing flaws. A parallel research literature, much smaller and almost entirely without public data, has shown that magnetometry can resolve precisely this missing information~\cite{ilott2018,romanenko2020,hu2020,ilott2017,nature2025magnetic,aachen2025}.

This work has accelerated significantly in the past several years. Optically-pumped magnetometers, nitrogen-vacancy diamond sensors, and SQUID arrays have matured into measurement systems suitable for routine use on commercial cells. The QuaLiProM consortium (Fraunhofer IFAM, FAU Erlangen, and industrial partners; BMBF-funded; running through November 2026) is explicitly building an industrial pipeline that combines magnetometry with deep learning for second-life classification of retired batteries~\cite{qualiprom}. The Aachen--J\"ulich--Sussex--PTB collaboration recently demonstrated quantum magnetic imaging of 6000\,mAh cells~\cite{aachen2025}. Operando magnetic microscopy of ionic and electronic current distributions appeared in \emph{Nature Communications} in late 2025~\cite{nature2025magnetic}. The trajectory is clear: magnetic sensing is becoming a first-class modality for battery diagnostics.

\subsection{The gap}

Despite this momentum, two things are conspicuously missing from the public landscape. First, \emph{to the best of our knowledge, no public dataset pairs battery magnetic measurements with degradation labels}. A systematic search of Zenodo, OSF, GitHub, Hugging Face, and the academic literature surfaces exactly one publicly archived raw magnetic-scan dataset of batteries: the OSF archive associated with Mohammadi, Ilott, and Jerschow (henceforth ``OSF'')~\cite{hu2020,osfdata}. It contains high-resolution magnetic field measurements of a single Li-ion cell at five operating voltages, with 41 scan positions per voltage. It does not include SOH labels, multi-cell variation, or a degradation trajectory. Every other publication reporting magnetic measurements of batteries uses proprietary or unreleased data.

Second, \emph{no bridge connects the rich electrochemical-degradation datasets to any magnetic-sensing modality}. Researchers building methods for magnetic SOH estimation, second-life classification, or anomaly detection have no public benchmark to develop against. Cross-lab comparison is essentially impossible.

This work asks: can the community make practical progress without waiting for paired magnetic--electrochemical data to become public, by \emph{bridging} the two modalities synthetically?

\section{The MagBridge-Battery v1.0 Dataset}
\label{sec:dataset}

MagBridge-Battery v1.0 is the central artifact of this work. We describe its composition, schema, splits, and integrity properties here. The bridge architecture that produced it is described in \S\ref{sec:architecture}; validation in \S\ref{sec:validation}.

\subsection{Composition}
\label{sec:dataset:composition}

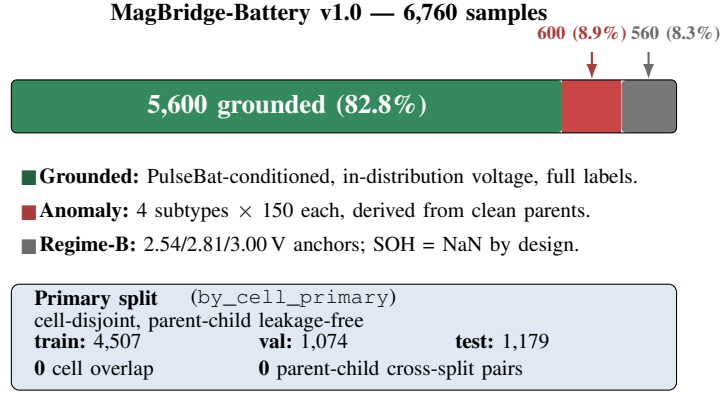
\begin{figure*}[!htbp]
\centering
\begin{tikzpicture}[font=\footnotesize,scale=1.0,transform shape]


  \node[anchor=south,font=\bfseries\small] at (4.2,3.75)
    {MagBridge-Battery v1.0 --- 6{,}760 samples};

  \fill[bridgegreen,rounded corners=2pt] (0,2.45) rectangle (7.29,3.15);
  \fill[anomred]                         (7.29,2.45) rectangle (8.07,3.15);
  \fill[regiongray,rounded corners=2pt] (8.07,2.45) rectangle (8.80,3.15);
  \draw[rounded corners=2pt] (0,2.45) rectangle (8.80,3.15);

  \node[white,font=\bfseries] at (3.65,2.80)
    {5{,}600 grounded (82.8\%)};

  \draw[-{Latex[length=2mm]},anomred!85!black,thick]
    (7.68,3.52) -- (7.68,3.16);
  \node[anchor=south,font=\scriptsize\bfseries,anomred!85!black]
    at (7.55,3.52) {600 (8.9\%)};

  \draw[-{Latex[length=2mm]},regiongray!90!black,thick]
    (8.43,3.52) -- (8.43,3.16);
  \node[anchor=south,font=\scriptsize\bfseries,regiongray!90!black]
    at (8.79,3.52) {560 (8.3\%)};

  \node[anchor=west,bridgegreen!70!black] at (0,1.85) {$\blacksquare$};
  \node[anchor=west] at (0.25,1.85)
    {\textbf{Grounded:} PulseBat-conditioned, in-distribution voltage, full labels.};

  \node[anchor=west,anomred!85!black] at (0,1.40) {$\blacksquare$};
  \node[anchor=west] at (0.25,1.40)
    {\textbf{Anomaly:} 4 subtypes $\times$ 150 each, derived from clean parents.};

  \node[anchor=west,regiongray!90!black] at (0,0.95) {$\blacksquare$};
  \node[anchor=west] at (0.25,0.95)
    {\textbf{Regime-B:} 2.54/2.81/3.00\,V anchors; SOH = NaN by design.};

\draw[
  rounded corners=3pt,
  fill=osfbluepale,
  draw=black
] (0,-0.95) rectangle (8.80,0.45);

\node[anchor=west,font=\footnotesize\bfseries] at (0.18,0.25)
  {Primary split};
\node[anchor=west,font=\footnotesize] at (2.25,0.27)
  {(\texttt{by\_cell\_primary})};

\node[anchor=west,font=\footnotesize] at (0.18,-0.05)
  {cell-disjoint, parent-child leakage-free};

\node[anchor=west,font=\footnotesize] at (0.18,-0.32)
  {\textbf{train:} 4{,}507};
\node[anchor=west,font=\footnotesize] at (3.15,-0.32)
  {\textbf{val:} 1{,}074};
\node[anchor=west,font=\footnotesize] at (5.75,-0.32)
  {\textbf{test:} 1{,}179};

\node[anchor=west,font=\footnotesize] at (0.18,-0.68)
  {\textbf{0} cell overlap};
\node[anchor=west,font=\footnotesize] at (3.15,-0.68)
  {\textbf{0} parent-child cross-split pairs};

\end{tikzpicture}
\caption{MagBridge-Battery v1.0 composition. The dataset contains 6{,}760 samples: 5{,}600 grounded samples, 600 synthetic anomaly samples, and 560 low-voltage Regime-B samples. Bar width is proportional to sample count. The primary benchmark split is \texttt{by\_cell\_primary}, which is cell-disjoint and parent-child leakage-free.}
\label{fig:composition}
\end{figure*}

The release contains \textbf{6{,}760 magnetic signatures} partitioned by provenance into three groups (Fig.~\ref{fig:composition}):

\begin{itemize}
    \item \textbf{5{,}600 PulseBat-conditioned grounded samples.} Generated by the bridge in the grounded regime ($v \in [3.06, 3.34]$\,V), conditioned on PulseBat-derived (SOH, SOC, U-features) drawn from real retired-cell pulse tests. Every sample carries the full label set.
    \item \textbf{600 synthetic sensor-anomaly samples.} Four subtypes, 150 each, derived from clean parents via controlled perturbations: \texttt{sensor\_dropout}, \texttt{calibration\_drift}, \texttt{temporal\_warp}, \texttt{periodic\_interference}. Each anomaly row carries a \texttt{parent\_sample\_id} pointing to its clean parent.
    \item \textbf{560 low-voltage Regime-B extrapolation samples.} Clustered at three low-voltage anchors (\texttt{nearest\_anchor} $\in \{2.54, 2.81, 3.00\}$\,V), outside the PulseBat in-distribution range. \emph{Regime-B samples are intended for low-voltage / OOD / anomaly-style evaluation, not SOH regression.} \texttt{soh}, \texttt{u\_features}, and \texttt{second\_life\_class} are \texttt{NaN} by design.
\end{itemize}

Every sample is a length-100 sequence with six signal channels plus a constant normalised time axis. MagBridge-Battery v1.0 uses the LFP subset of PulseBat exclusively; NMC and LMO records in PulseBat are reserved for future cross-chemistry extensions (\S\ref{sec:limitations}).

\subsection{Schema}
\label{sec:dataset:schema}

Each row carries six signal channels of length 100:
\begin{center}
\texttt{B\_s1Y}, \texttt{B\_s1Z}, \texttt{B\_s2Y}, \texttt{B\_s2Z}, \texttt{B\_s1C5}, \texttt{B\_s2C6}.
\end{center}
The first four are signed Y/Z components of sensors 1 and 2. The last two are the channel-5 and channel-6 fields from the OSF source; the OSF archive labels these channels \texttt{Mag}, but their values are signed and can legitimately be negative (123 rows contain negative entries in \texttt{B\_s1C5}; 86 in \texttt{B\_s2C6}; minima $-80.47$ and $-94.08$ respectively). We rename them \texttt{C5} and \texttt{C6} in the release schema to avoid implying a strict $\sqrt{Y^2 + Z^2}$ magnitude interpretation. \textbf{\texttt{B\_s1C5} and \texttt{B\_s2C6} are retained signed source channels and are not interpreted as strict physical magnitudes.}

A \texttt{time\_norm} column carries 100 evenly-spaced values in $[0,1]$. It is the same vector for every sample --- a constant reference grid included for loader convenience, droppable without information loss. The \texttt{temporal\_warp} anomaly perturbs signal values on this fixed grid; it does not export an irregular per-sample timebase.

Metadata fields include identifiers (\texttt{sample\_id}, \texttt{parent\_sample\_id}, \texttt{cell\_id}, \texttt{generation\_seed}), provenance (\texttt{bridge\_version}, \texttt{bridge\_config\_hash}, \texttt{schema\_version}), state labels (\texttt{voltage}, \texttt{soc}, \texttt{soh}, \texttt{chemistry}, \texttt{regime}, \texttt{nearest\_anchor}, \texttt{u\_features}, \texttt{second\_life\_class}), and anomaly labels (\texttt{anomaly\_flag}, \texttt{anomaly\_subtype}, \texttt{anomaly\_origin}, \texttt{anomaly\_severity}).

\subsection{Benchmark splits}
\label{sec:dataset:splits}

We provide two splits.

\textbf{All primary benchmark results are reported on the \texttt{by\_cell\_primary} split}, which is cell-disjoint and parent-child leakage-free. Train / validation / test counts are 4{,}507 / 1{,}074 / 1{,}179 (Fig.~\ref{fig:composition}). Verified guarantees: zero physical cells overlap between subsets, zero (clean parent, anomaly child) pairs cross subset boundaries, zero sample-ID overlap. This is the split to use for any reported number.

The companion \texttt{by\_record\_optimistic\_baseline} split is provided \emph{only} as a contrast and is not recommended for benchmark reporting. Its leakage is quantified explicitly: 59 cells appear in more than one subset, and 292 parent-child pairs are split across subset boundaries. A model trained on this split will appear artificially strong; we ship it so the inflation effect can be measured directly.

\subsection{Integrity properties}

The release was audited on the shipped artifacts. Verified: zero duplicate \texttt{sample\_id} values across all 6{,}760 rows; zero duplicate full-signal hashes; zero NaN or infinity entries in any of the six signal channels; uniform length-100 signal arrays; exact metadata-to-shard ID correspondence; valid parent-child links for all 600 synthetic anomalies (all parents exist and are clean); and zero residual occurrences of deprecated schema fields or legacy anomaly labels. SHA-256 checksums for every shipped file are included in the release bundle.

\subsection{File layout, license, and citation}
\label{sec:dataset:release}

The release ships as a single bundle: five Parquet shards of 1{,}352 rows each, a metadata-only Parquet view, the two split JSONs, a generation manifest pinning bridge version and configuration hash, a minimal Python loader, SHA-256 checksums, and licensing files. The dataset is licensed CC-BY-4.0; release code is licensed Apache-2.0; the LICENSE file documents the upstream sources (OSF magnetometry archive and PulseBat dataset) and their respective license declarations. \emph{No raw upstream data is redistributed in this release}; aggregate statistics derived from the OSF archive (per-anchor means and variances) are embedded in the bridge implementation but not in the released data files.

Users are kindly requested to cite both this paper and the dataset DOI; see \texttt{CITING.md} in the release bundle for the recommended dual citation. The release is on Zenodo (DOI: \texttt{10.5281/zenodo.20260147}) and the code is on GitHub at \url{https://github.com/SakthiGs/MagBridge-Battery}.

\section{Bridge Architecture}
\label{sec:architecture}

The bridge is a deterministic function $\mathcal{B}(v, \text{SOC}, \text{SOH}, \mathbf{u}; \theta) \rightarrow \mathbf{X} \in \mathbb{R}^{T \times C}$ that maps a generation request --- operating voltage $v$, state of charge SOC, state of health SOH, and PulseBat U-feature vector $\mathbf{u} \in \mathbb{R}^{21}$ --- to a synthetic magnetic-signature time series of length $T = 100$ across $C = 6$ channels (\S\ref{sec:dataset:schema}). The configuration $\theta$ collects all tunable parameters, fixed once at bridge instantiation. Generation is reproducible: a given (request, $\theta$, seed) tuple always produces the same output.

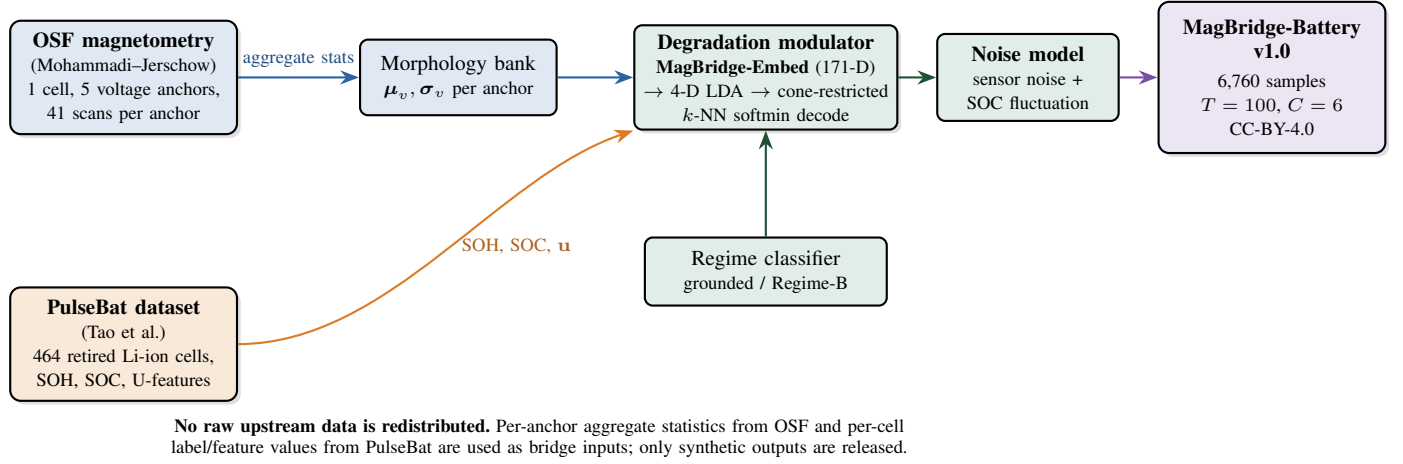
\begin{figure*}[!htbp]
\centering
\begin{tikzpicture}[font=\footnotesize,node distance=1.4cm and 0.9cm,>={Stealth[length=2.5mm]}]

  \node[draw,thick,rounded corners=4pt,fill=osfbluepale,
        minimum width=3.0cm,minimum height=1.5cm,align=center,
        drop shadow={shadow xshift=0.5mm,shadow yshift=-0.5mm,opacity=0.25}]
        (osf) {\textbf{OSF magnetometry}\\\scriptsize{(Mohammadi--Jerschow)}\\\scriptsize{1 cell, 5 voltage anchors,}\\\scriptsize{41 scans per anchor}};

  \node[draw,thick,rounded corners=4pt,fill=pulseorangepale,below=2.0cm of osf,
        minimum width=3.0cm,minimum height=1.5cm,align=center,
        drop shadow={shadow xshift=0.5mm,shadow yshift=-0.5mm,opacity=0.25}]
        (pulse) {\textbf{PulseBat dataset}\\\scriptsize{(Tao et al.)}\\\scriptsize{464 retired Li-ion cells,}\\\scriptsize{SOH, SOC, U-features}};

  \node[draw,thick,rounded corners=3pt,fill=osfbluepale,right=1.6cm of osf,
        minimum width=2.6cm,minimum height=1.0cm,align=center]
        (morph) {Morphology bank\\\scriptsize{$\boldsymbol{\mu}_v,\boldsymbol{\sigma}_v$ per anchor}};

  \node[draw,thick,rounded corners=3pt,fill=bridgegreenpale,right=1.0cm of morph,
        minimum width=3.2cm,minimum height=1.3cm,align=center,
        drop shadow={shadow xshift=0.5mm,shadow yshift=-0.5mm,opacity=0.25}]
        (mod) {\textbf{Degradation modulator}\\\scriptsize{\textbf{MagBridge-Embed} (171-D)}\\\scriptsize{$\rightarrow$ 4-D LDA $\rightarrow$ cone-restricted}\\\scriptsize{$k$-NN softmin decode}};

  \node[draw,thick,rounded corners=3pt,fill=bridgegreenpale,below=1.4cm of mod,
        minimum width=3.2cm,minimum height=0.9cm,align=center]
        (regime) {Regime classifier\\\scriptsize{grounded / Regime-B}};

  \node[draw,thick,rounded corners=3pt,fill=bridgegreenpale,right=0.5cm of mod,
      minimum width=2.4cm,minimum height=1.15cm,align=center]
      (noise) {\textbf{Noise model}\\
      \scriptsize sensor noise +\\
      \scriptsize SOC fluctuation};

  \node[draw,thick,rounded corners=4pt,fill=outpurplepale,right=0.5cm of noise,
        minimum width=3.0cm,minimum height=2.0cm,align=center,
        drop shadow={shadow xshift=0.5mm,shadow yshift=-0.5mm,opacity=0.25}]
        (out) {\textbf{MagBridge-Battery}\\\textbf{v1.0}\\[2pt]
              \scriptsize{6{,}760 samples}\\
              \scriptsize{$T=100$, $C=6$}\\
              \scriptsize{CC-BY-4.0}};

  \draw[->,thick,osfblue] (osf) -- node[above,midway,font=\scriptsize,osfblue!80!black]{aggregate stats} (morph);
  \draw[->,thick,osfblue] (morph) -- (mod);
  \draw[->,thick,pulseorange] (pulse.east) .. controls +(2,0) and +(-1.8,-0.6) .. node[above,pos=0.7,yshift=-25pt,font=\scriptsize,pulseorange!80!black]{SOH, SOC, $\mathbf{u}$} (mod.south west);
  \draw[->,thick,bridgegreen!60!black] (regime) -- (mod);
  \draw[->,thick,bridgegreen!60!black] (mod) -- (noise);
  \draw[->,thick,outpurple] (noise) -- (out);

  \node[anchor=north,font=\scriptsize,align=center,text width=15cm] at (5.5,-4.4)
  {\textbf{No raw upstream data is redistributed.} Per-anchor aggregate statistics from OSF and per-cell label/feature values from PulseBat are used as bridge inputs; only synthetic outputs are released.};

\end{tikzpicture}
\caption{Bridge architecture. Real magnetic morphology from the OSF archive and SOH/SOC/U-feature labels from PulseBat are combined through a morphology bank, MagBridge-Embed degradation modulator, regime classifier, and noise model to generate MagBridge-Battery v1.0. Only synthetic outputs are released; no raw upstream files are redistributed.}
\label{fig:architecture}
\end{figure*}

Figure~\ref{fig:architecture} summarises the bridge's data flow. The bridge has four components, applied in sequence: a regime classifier, a morphology bank derived from OSF, a degradation modulator conditioned on PulseBat labels via a learned latent representation we call \textbf{MagBridge-Embed}, and a noise model. The full mathematical specification of the degradation modulator (equations for LDA projection, perturbation, $k$-NN softmin decoding, and base--modulated blending) is provided in Appendix~\ref{app:archmath}; we describe the architecture at the conceptual level here.

\subsection{Regime classifier}

The bridge handles two operating regimes, derived from cross-data analysis of the OSF voltage anchors and the PulseBat U-feature distribution:

\begin{itemize}
    \item \textbf{Grounded regime} ($v \in [3.06, 3.34]$\,V): both OSF morphology and PulseBat conditioning populate this range. The bridge interpolates between the OSF anchors at 3.10\,V and 3.34\,V using PulseBat-derived (SOH, SOC) as conditioning.
    \item \textbf{Regime~B (extrapolation)} ($v \in [2.54, 3.06)$\,V): only OSF populates this range; PulseBat does not test in over-discharge for safety reasons. The bridge reproduces OSF morphology here without PulseBat-grounded conditioning. As stated in \S\ref{sec:dataset:composition}, Regime-B samples are intended for low-voltage / OOD evaluation, not SOH regression.
\end{itemize}

Voltages outside $[2.54, 3.34]$\,V are unsupported and rejected. Every generated sample carries its regime as metadata.

\subsection{Morphology bank}

The OSF archive is canonicalised into per-anchor empirical statistics: at each of the five anchor voltages $\{2.54, 2.81, 3.00, 3.10, 3.34\}$\,V, the bridge extracts the mean trajectory $\boldsymbol{\mu}_v \in \mathbb{R}^{T \times C}$ and per-timestep, per-channel standard deviation $\boldsymbol{\sigma}_v \in \mathbb{R}^{T \times C}$ across all 41 scan positions. For a generation request at voltage $v$, the bank produces a base sample by linear interpolation between bracketing anchors (full equations in Appendix~\ref{app:archmath}). At an exact OSF anchor with no perturbation, the bank reproduces the empirical mean trajectory to machine precision (sanity invariant 1, \S\ref{sec:validation}).

\subsection{Degradation modulator and MagBridge-Embed}
\label{sec:arch:modulator}

The degradation modulator applies SOH-driven perturbation to the base morphology through a learned latent representation. We refer to this representation as \textbf{MagBridge-Embed}: a 171-dimensional embedding produced by a fixed quantum reservoir computer~\cite{fujii2017qrc} (10 qubits arranged as 4 memory + 6 processor, 2 reservoir layers, pooling $\{$last, mean, std$\}$). The reservoir parameters are not trained --- only the downstream linear readouts that operate in this embedding space are.

The modulator's data flow is:
\begin{enumerate}
\item Compute the MagBridge-Embed of the base sample.
\item Project to a 4-D Linear Discriminant Analysis (LDA) subspace fit on real OSF samples with voltage anchors as classes. In this subspace, within-anchor scatter is $\sim 1.78$ and inter-anchor separation is $\sim 279$ (ratio $\sim 157$).
\item Perturb in the LDA subspace along the fitted state direction with magnitude proportional to $(1 - \text{SOH})$.
\item Decode back to the time domain via cone-restricted $k$-NN softmin retrieval against the 205 OSF samples, with $k=8$ and a $75^\circ$ alignment cone.
\item Blend the decoded morphology with the base sample proportionally to $(1 - \text{SOH})$.
\item Apply per-channel amplitude scaling and SOH-scaled spectral broadening; the spectral broadening is mildly modulated by PulseBat U-feature dispersion.
\end{enumerate}

Section~\ref{sec:results:ablation} reports what the validation suite reveals about which of these components are load-bearing for downstream SOH decoding.

\subsection{Noise model}

Two additive noise sources are applied to the modulator output: \emph{sensor noise} (zero-mean Gaussian, per-channel standard deviation set to 5\% of the corresponding OSF anchor's empirical channel std), and \emph{SOC-dependent fluctuation} (low-frequency component scaled by $|\text{SOC} - 50|/50$, implemented as boxcar-smoothed Gaussian noise normalised and scaled to 4\% of per-channel range). Ageing-induced disorder is intentionally not added at this stage to avoid double-counting with the spectral broadening already applied.

\subsection{Synthetic sensor anomalies}

For the 600 anomaly samples in the v1.0 release, a clean parent sample is generated by the pipeline above, and one of four perturbations is applied: \texttt{sensor\_dropout}, \texttt{calibration\_drift}, \texttt{temporal\_warp}, or \texttt{periodic\_interference}. The clean parent's \texttt{sample\_id} is preserved as \texttt{parent\_sample\_id} on the anomaly row. The primary split is constructed so that no parent-child pair is split across train/val/test boundaries.

\section{Validation methodology}
\label{sec:validation}

A bridge that generates plausible-looking samples is not enough. We validate the dataset in three nested layers, each of which the bridge must pass before its outputs are released.

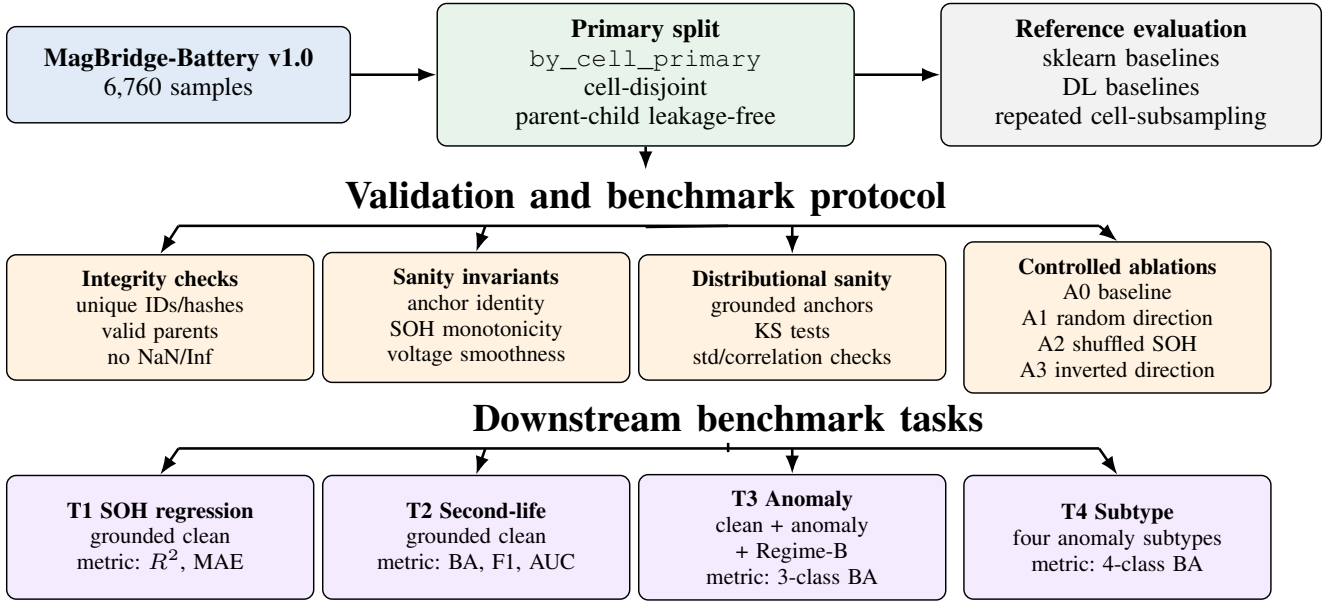
\begin{figure*}[!t]
\centering
\resizebox{0.96\textwidth}{!}{%
\begin{tikzpicture}[
  font=\footnotesize,
  >=Latex,
  box/.style={
    draw,
    rounded corners=3pt,
    align=center,
    inner sep=5pt,
    line width=0.45pt
  },
  topbox/.style={
    box,
    fill=mbblue,
    text width=3.4cm,
    minimum height=1.05cm,
    font=\footnotesize
  },
  midbox/.style={
    box,
    fill=mbgreen,
    text width=4.2cm,
    minimum height=1.25cm,
    font=\footnotesize
  },
  refbox/.style={
    box,
    fill=mbgray,
    text width=3.8cm,
    minimum height=1.25cm,
    font=\footnotesize
  },
  checkbox/.style={
    box,
    fill=mborange,
    text width=3.0cm,
    minimum height=1.35cm,
    font=\scriptsize
  },
  taskbox/.style={
    box,
    fill=mbpurple,
    text width=3.0cm,
    minimum height=1.35cm,
    font=\scriptsize
  },
  arrow/.style={
    -{Latex[length=2mm]},
    thick
  }
]

\node[topbox] (dataset) at (-1.6,4.0)
  {\textbf{MagBridge-Battery v1.0}\\6{,}760 samples};

\node[midbox] (split) at (3.5,4.0)
  {\textbf{Primary split}\\
   \texttt{by\_cell\_primary}\\
   cell-disjoint\\
   parent-child leakage-free};

\node[refbox] (eval) at (8.8,4.0)
  {\textbf{Reference evaluation}\\
   sklearn baselines\\
   DL baselines\\
   repeated cell-subsampling};

\draw[arrow] (dataset.east) -- (split.west);
\draw[arrow] (split.east) -- (eval.west);

\node[font=\bfseries\large] (protocol) at (3.5,2.65)
  {Validation and benchmark protocol};

\draw[arrow] (split.south) -- (protocol.north);

\node[checkbox] (integrity) at (-1.8,1.35)
  {\textbf{Integrity checks}\\
   unique IDs/hashes\\
   valid parents\\
   no NaN/Inf};

\node[checkbox] (sanity) at (1.65,1.35)
  {\textbf{Sanity invariants}\\
   anchor identity\\
   SOH monotonicity\\
   voltage smoothness};

\node[checkbox] (dist) at (5.10,1.35)
  {\textbf{Distributional sanity}\\
   grounded anchors\\
   KS tests\\
   std/correlation checks};

\node[checkbox] (abl) at (8.65,1.35)
  {\textbf{Controlled ablations}\\
   A0 baseline\\
   A1 random direction\\
   A2 shuffled SOH\\
   A3 inverted direction};

\draw[thick] (protocol.south) -- (4.4,2.35);
\draw[thick] (-1.6,2.35) -- (8.45,2.35);
\draw[arrow] (-1.6,2.35) -- (integrity.north);
\draw[arrow] (1.75,2.35) -- (sanity.north);
\draw[arrow] (5.10,2.35) -- (dist.north);
\draw[arrow] (8.45,2.35) -- (abl.north);

\node[font=\bfseries\large] (taskslabel) at (4.4,0.25)
  {Downstream benchmark tasks};

\node[taskbox] (t1) at (-1.8,-1.05)
  {\textbf{T1 SOH regression}\\
   grounded clean\\
   metric: $R^2$, MAE};

\node[taskbox] (t2) at (1.65,-1.05)
  {\textbf{T2 Second-life}\\
   grounded clean\\
   metric: BA, F1, AUC};

\node[taskbox] (t3) at (5.10,-1.05)
  {\textbf{T3 Anomaly}\\
   clean + anomaly\\
   + Regime-B\\
   metric: 3-class BA};

\node[taskbox] (t4) at (8.65,-1.05)
  {\textbf{T4 Subtype}\\
   four anomaly subtypes\\
   metric: 4-class BA};

\draw[thick] (taskslabel.south) -- (4.4,-0.12);
\draw[thick] (-1.6,-0.08) -- (8.45,-0.08);
\draw[arrow] (-1.6,-0.08) -- (t1.north);
\draw[arrow] (1.75,-0.08) -- (t2.north);
\draw[arrow] (5.10,-0.08) -- (t3.north);
\draw[arrow] (8.45,-0.08) -- (t4.north);

\end{tikzpicture}%
}

\caption{Benchmark and validation protocol for MagBridge-Battery v1.0. The release is evaluated through integrity checks, bridge sanity invariants, distributional sanity at grounded OSF anchors, controlled ablations, and four downstream benchmark tasks on the \texttt{by\_cell\_primary} leakage-safe split. BA denotes balanced accuracy.}
\label{fig:validation_protocol}
\end{figure*}

Figure~\ref{fig:validation_protocol} summarises the validation and benchmark protocol used for MagBridge-Battery v1.0, including structural checks, falsification ablations, and downstream tasks evaluated on the leakage-safe primary split.

\subsection{Sanity invariants}

Five structural invariants are tested automatically on every release candidate; failure on any blocks release.

\begin{enumerate}
    \item \textbf{Identity at OSF anchors.} At each of the five OSF anchor voltages, the bridge's deterministic anchor-replica function (no perturbation, no noise) reproduces the empirical mean trajectory to within $10^{-12}$ of machine precision.
    \item \textbf{SOH monotonicity.} For fixed (voltage, SOC, seed), increasing $(1 - \text{SOH})$ from 0 to 0.25 produces monotonically increasing L2 distance from the SOH$=$1.0 reference in at least one defensible signature metric.
    \item \textbf{Voltage smoothness.} A 0.01\,V perturbation in voltage produces a smaller change in output than a 0.10\,V perturbation, ruling out discontinuities at anchor boundaries.
    \item \textbf{Regime classification correctness.} The classifier returns the expected regime at boundary cases $\{2.54, 2.81, 3.00, 3.06, 3.10, 3.34\}$\,V and rejects out-of-range tests at $\{2.00, 4.00\}$\,V.
    \item \textbf{Anomaly flag consistency.} Every Regime-B sample carries \texttt{anomaly\_flag} $=$ \texttt{True}; every grounded-regime clean sample carries \texttt{False}.
\end{enumerate}

\subsection{Distributional sanity at grounded anchors}

At the two grounded-regime OSF anchors (3.10\,V, 3.34\,V), bridge-generated samples at SOH$=$1.0 should be statistically indistinguishable from real OSF samples. We generate 41 synthetic samples per anchor (matching the OSF scan-position count) and run per-channel, per-timestep two-sample Kolmogorov--Smirnov tests against real OSF samples. We additionally report (a) the ratio of synthetic to real per-channel standard deviations as an amplitude calibration check, and (b) the mean absolute difference of the synthetic and real cross-channel correlation matrices.

\subsection{Controlled ablations}
\label{sec:validation:ablations}

A bridge that passes the invariants and the distributional sanity check could still be a label-aligned artifact --- it could encode the SOH input into the output in a way that any standard readout trivially recovers, without the bridge doing physically meaningful work. To rule this out, we run four ablation scenarios on identical request lists:

\begin{itemize}
    \item \textbf{A0 baseline}: real fitted state direction, real per-cell SOH labels.
    \item \textbf{A1 random direction}: replace the fitted state direction with a uniformly-sampled unit vector in the 4-D LDA subspace.
    \item \textbf{A2 shuffled SOH labels}: permute SOH values across PulseBat records before the bridge sees them, breaking the SOH--cell binding. The benchmark grades regression against the \emph{original} per-cell SOH (preserved in metadata), not the shuffled values driving generation.
    \item \textbf{A3 inverted direction}: replace the fitted state direction with its negation.
\end{itemize}

A2 is the load-bearing test. If the bridge is encoding cell-state-truthful information, A2 should collapse downstream regression $R^2$ to approximately zero. A1 and A3 jointly probe whether the specific fitted LDA direction is privileged for downstream decoding.

\subsection{Relationship of ablation pilots to the v1.0 release}
\label{sec:validation:pilot_vs_release}

The four-scenario ablations were executed on focused $\sim$310-sample-per-scenario pilots produced by the v1.0 bridge architecture in the grounded regime. The released v1.0 dataset (\S\ref{sec:dataset}) is generated by the \emph{same} v1.0 bridge with the \emph{same} configuration $\theta$ (pinned by \texttt{bridge\_config\_hash} in the manifest), scaled to 5{,}600 grounded samples plus the 600 anomalies and 560 Regime-B samples. Because the released dataset is generated by the same bridge configuration, the ablation pilots provide architecture-level evidence for the released dataset's behaviour; we do not re-run ablations at full scale because direction-perturbation contrasts are already statistically saturated at the pilot size.

\subsection{Benchmark tasks}
\label{sec:validation:tasks}

For downstream evaluation we define three tasks:

\begin{itemize}
    \item \textbf{T1 SOH regression}: continuous SOH (released range $[0.744, 0.962]$, reflecting PulseBat retired-cell SOH distribution) from the magnetic signature. Grounded-regime samples only.
    \item \textbf{T2 Second-life classification}: binary classification with the cutoff at SOH$=0.85$ (\texttt{reuse} if SOH$>0.85$, \texttt{recondition} otherwise), matching the convention used in the released metadata's \texttt{second\_life\_class} field. Grounded-regime samples only.
    \item \textbf{T3 Anomaly detection}: train on grounded-regime clean samples, test on a mix of held-out grounded clean, synthetic-anomaly perturbed, and Regime-B samples.
\end{itemize}

All three tasks use the \texttt{by\_cell\_primary} split (\S\ref{sec:dataset:splits}) with five seeds, and few-shot stratified subsampling at $k \in \{2, 5, 10, 20\}$ examples per class (T1, T2) or examples per training pool (T3). Reference models are standard sklearn pipelines: ridge regression, SVR-RBF, and random forest for T1; logistic regression, ridge classifier, linear SVC, and random forest for T2 and T3. The benchmark suite uses a 57-feature static descriptor (9 features per channel plus 3 cross-channel correlations), which reproduces a reference OSF feature implementation to $1.87 \times 10^{-12}$ machine precision, ensuring that any difference between bridge results and a reference experiment on real OSF data is attributable to the bridge itself, not feature drift.

\section{Results}
\label{sec:results}

\subsection{Sanity invariants}

All five sanity invariants pass. Identity at OSF anchors: maximum absolute deviation $1.87 \times 10^{-12}$. SOH monotonicity: distance from SOH$=$1.0 reference grows monotonically across $\text{SOH} \in \{1.00, 0.95, 0.90, 0.85, 0.80, 0.75\}$ at both 3.10\,V and 2.81\,V. Voltage smoothness: a 0.01\,V step produces L2 change $\approx 261$ vs.\ $\approx 1881$ for a 0.10\,V step. Regime classification and anomaly flag invariants pass on all boundary cases.

\subsection{Distributional sanity at grounded anchors}

Per-channel KS tests pass on 6 of 6 channels at both grounded anchors (3.10\,V, 3.34\,V) at $p > 0.05$. Per-channel standard-deviation ratios (synthetic / real) lie in $[0.98, 1.02]$, indicating essentially exact amplitude calibration. The mean absolute cross-channel correlation difference is 0.46 at 3.10\,V and 0.35 at 3.34\,V; the v1 bridge's independent-channel-perturbation assumption is the source of this gap and is a known limitation (\S\ref{sec:limitations}). Qualitatively, bridge-generated samples at SOH$=$1.0 overlay almost exactly onto real OSF samples at both grounded anchors across all six channels; the bridge captures the characteristic dip-spike feature near timestep $\approx 45$ in channels \texttt{B\_s1C5} and \texttt{B\_s2C6}, and per-anchor amplitude relationships across the six channels are preserved.

\subsection{Ablation results}
\label{sec:results:ablation}

Headline results from the controlled ablations are summarised in Tables~\ref{tab:soh_ablation} and~\ref{tab:second_life_ablation}.

\begin{table}[t]
\caption{SOH regression $R^2$ (best model per shot, by-cell test split, 5-seed mean) on bridge v1.0 ablation pilots.}
\label{tab:soh_ablation}
\centering
\begin{tabular}{lcccc}
\toprule
Scenario & $k=2$ & $k=5$ & $k=10$ & $k=20$ \\
\midrule
A0 baseline & $+0.62$ & $+0.64$ & $+0.70$ & $+0.77$ \\
A1 random direction & $+0.67$ & $+0.61$ & $+0.72$ & $+0.78$ \\
A2 shuffled SOH & $-0.01$ & $+0.03$ & $-0.00$ & $-0.04$ \\
A3 inverted direction & $+0.67$ & $+0.62$ & $+0.72$ & $+0.75$ \\
\bottomrule
\end{tabular}
\end{table}

\begin{table}[t]
\caption{Second-life classification balanced accuracy (best model per shot, by-cell test split, 5-seed mean). Chance is 0.50.}
\label{tab:second_life_ablation}
\centering
\begin{tabular}{lcccc}
\toprule
Scenario & $k=2$ & $k=5$ & $k=10$ & $k=20$ \\
\midrule
A0 baseline & $0.72$ & $0.84$ & $0.84$ & $0.93$ \\
A1 random direction & $0.76$ & $0.85$ & $0.85$ & $0.92$ \\
A2 shuffled SOH & $0.56$ & $0.61$ & $0.56$ & $0.56$ \\
A3 inverted direction & $0.74$ & $0.82$ & $0.85$ & $0.94$ \\
\bottomrule
\end{tabular}
\end{table}

Two patterns are visible.

\textbf{A2 (shuffled SOH) collapses cleanly.} SOH regression $R^2$ drops from $+0.77$ to $-0.04$ at $k=20$, a swing of more than 0.8. Second-life accuracy falls to chance across all shot levels. Both tasks confirm that the bridge encodes input SOH non-trivially: when the SOH--cell binding is broken at generation time, downstream readouts cannot recover SOH from the synthetic signatures. This is the dataset's primary validation: it is not producing label-aligned artefacts.

\textbf{A1 and A3 perform almost identically to A0.} Across all four shot levels and both tasks, random and inverted perturbation directions match the principled-direction baseline within statistical noise (the largest cross-scenario gap is 0.05 $R^2$). We address what this means in \S\ref{sec:discussion:direction}.

\subsection{Full-release benchmark on \texttt{by\_cell\_primary}}
\label{sec:results:fullbench}

The ablation pilots of \S\ref{sec:results:ablation} characterise the bridge's response to controlled perturbations. We additionally report headline numbers on the full released v1.0 dataset under the primary leakage-safe split (Table~\ref{tab:fullbench}), using standard sklearn pipelines and reporting the best model per task across five repeated cell-subsampling seeds (each seed sees a different 80\% of the training cells). These are the numbers users should compare against when developing methods on MagBridge-Battery.

\begin{table}[!htbp]
\caption{Full MagBridge-Battery v1.0 benchmark on the \texttt{by\_cell\_primary} cell-disjoint split. Best model per task; 5 cell-subsampling seeds; mean $\pm$ std. Models considered: Ridge / SVR-RBF / RF for T1; LogReg / RidgeCls / LinSVC / RF for T2--T4.}
\label{tab:fullbench}
\centering
\begin{tabular}{llll}
\toprule
Task & Best model & Metric & Value \\
\midrule
T1 SOH regression              & Ridge    & $R^2$ & $+0.675 \pm 0.001$ \\
T2 Second-life classification  & RidgeCls & BA     & $0.907 \pm 0.001$ \\
T3 Anomaly detection (3-class) & RF       & BA     & $0.789 \pm 0.003$ \\
T4 Anomaly subtype (4-class)   & RF       & BA     & $0.725 \pm 0.028$ \\
\bottomrule
\end{tabular}

\vspace{2pt}
\footnotesize{\textit{Note:} BA denotes balanced accuracy.}
\end{table}

T4 (anomaly subtype classification --- distinguishing among the four subtypes \texttt{sensor\_dropout}, \texttt{calibration\_drift}, \texttt{temporal\_warp}, and \texttt{periodic\_interference}) is reported as an additional benchmark task. A random forest achieves $0.725 \pm 0.028$ on 4-way classification (chance $=0.25$), substantially above chance but with clear headroom for stronger readouts (sequence models, learned embeddings). T1 also leaves significant headroom (R$^2$ of $\approx 0.68$ with ridge regression on static features), suggesting that sequence-aware methods could meaningfully improve SOH regression. T2 is nearly saturated by classical baselines because second-life classification is essentially a thresholded view of SOH; once T1 is solved well, T2 follows. T3 sits in the middle, indicating that anomaly detection on this dataset is genuinely non-trivial under the cell-disjoint protocol but not intractable.

A note on the T3 anomaly-detection metric. T3 is a 3-way task under the released benchmark protocol: synthetic-anomaly samples are derived from clean parents and differ from them only by the perturbation type, while Regime-B samples differ from grounded-regime samples primarily by voltage. A simplified regime-separation sanity check using a random forest can reach $\text{AUROC} \approx 1.00$ when only the Regime-B-vs-grounded distinction is being decided, but we do not treat this as the primary benchmark, because it reflects strong voltage-anchor separation rather than discriminative learning of anomaly structure. The headline T3 balanced accuracy of $0.789$ in Table~\ref{tab:fullbench} is computed on the full 3-way protocol that requires distinguishing perturbed-grounded, clean-grounded, and Regime-B samples on the cell-disjoint test split.

\subsection{Deep-learning baselines}
\label{sec:results:dl}

We additionally evaluated three off-the-shelf neural baselines to quantify whether the raw length-100 signatures are immediately exploitable by standard sequence models: a 3-layer MLP on the flattened raw signal, a 3-block 1D CNN, and a single-layer LSTM. All three operate directly on the six-channel sequences after channel-wise standardization using training-set statistics. We trained with Adam (learning rate $10^{-3}$, weight decay $10^{-4}$, batch size 64) for up to 15 epochs with early stopping. We did not tune architecture depth, receptive field, or class weighting per individual task; these baselines are intended as reproducible off-the-shelf references rather than tuned ceilings. Results across three repeated cell-subsampling seeds are shown in Table~\ref{tab:dlbench}.

\begin{table*}[!t]
\caption{Off-the-shelf deep-learning baselines on the \texttt{by\_cell\_primary} cell-disjoint split. Results are mean $\pm$ standard deviation over three repeated cell-subsampling seeds. The best classical baseline from Table~\ref{tab:fullbench} is reproduced for comparison.}
\label{tab:dlbench}
\centering
\footnotesize
\begin{tabular}{lllll}
\toprule
Task & MLP & 1D-CNN & LSTM & Classical \\
\midrule
T1 SOH, $R^2$
& $-2.51 \pm 1.24$
& $\phantom{-}0.17 \pm 0.24$
& $-0.09 \pm 0.03$
& $0.675$ \\
T2 Second-life, BA
& $0.599 \pm 0.016$
& $\mathbf{0.858 \pm 0.056}$
& $0.833 \pm 0.053$
& $0.907$ \\
T3 Anomaly, BA
& $\mathbf{0.755 \pm 0.028}$
& $0.365 \pm 0.003$
& $0.672 \pm 0.006$
& $0.789$ \\
T4 Subtype, BA
& $\mathbf{0.355 \pm 0.024}$
& $0.340 \pm 0.024$
& $0.299 \pm 0.021$
& $0.725$ \\
\bottomrule
\end{tabular}

\vspace{2pt}
\footnotesize{\textit{Note:} BA denotes balanced accuracy. Bold indicates the best deep-learning baseline for each classification task.}
\end{table*}

The neural baselines establish a useful lower bound for direct sequence-learning methods rather than a tuned deep-learning ceiling. The best neural models approach the static-feature classical baselines on T2 and T3: the 1D CNN reaches $0.858$ balanced accuracy on second-life classification compared with $0.907$ for RidgeCls, and the MLP reaches $0.755$ balanced accuracy on the 3-class anomaly task compared with $0.789$ for the random forest. In contrast, the gap is substantial on T1 and T4: the best neural model reaches only $R^2 = 0.17$ on SOH regression compared with $0.675$ for ridge regression, and $0.355$ balanced accuracy on anomaly-subtype classification compared with $0.725$ for the random forest. The MLP shows unstable behaviour on T1 (negative $R^2$ with high variance), consistent with overfit on the flattened raw signal under limited training data.

We interpret this gap as a benchmark signal rather than a dataset deficiency. The 57-feature static descriptor used by the classical baselines encodes compact domain summaries --- channel means, variation, energy, slopes, and cross-channel correlations --- that are effective at this release size and split. Closing the gap likely requires representation-learning methods with stronger inductive bias, pretraining, contrastive objectives, or sequence encoders tuned specifically for magnetic-signature morphology. MagBridge-Battery is therefore not solved by off-the-shelf application of standard deep-learning architectures, while leaving clear headroom for stronger learned representations.

\section{Discussion}
\label{sec:discussion}

\subsection{What the A2 collapse establishes}

When PulseBat SOH labels are randomly permuted across cells before the bridge sees them, downstream SOH regression collapses from $R^2 \approx 0.77$ to $\approx 0$. This makes the alternative explanations --- bridge-generated samples being SOH-decodable through correlated covariates, dataset leakage, or label memorisation --- unlikely under the tested protocol. In practical terms, a researcher training an SOH estimator on MagBridge-Battery is using bridge-consistent SOH-conditioned signal rather than a purely label-aligned artefact.

\subsection{What A1$\approx$A0 means}
\label{sec:discussion:direction}

The data show that random and inverted perturbation directions in the 4-D LDA subspace produce SOH-decoding accuracy within $\sim 0.05$ $R^2$ of the principled fitted direction. The architectural reason is geometric: the OSF dataset contains exactly five anchor clusters in the 4-D subspace, separated by $\sim 280$ units with within-cluster scatter $\sim 1.8$ (a separation ratio of 157). Any perturbation of meaningful magnitude moves the embedding into a different anchor's neighbourhood, regardless of direction. The downstream SOH-decoding signal is generated by the resulting blend, with amplitude scaling proportional to $(1 - \text{SOH})$.

In other words, the bridge produces decodable SOH content via \emph{anchor blending under SOH-driven blend amplitude} --- a property inherited from the bank's small, sharply-clustered geometry, not from any specific embedding direction. We treat this as a property of retrieve-and-blend bridge architectures over small morphology banks, of which MagBridge-Battery is one. A more flexible decoder (e.g.\ a conditional generative model trained on OSF data) would in principle let the conditioning direction shape output distributions non-trivially, and would change this finding; we treat this as the most promising single-axis future-work direction (\S\ref{sec:limitations}).

\subsection{Implications for synthetic-data work in this field}

Two implications follow. First, bridge architectures that retrieve from a finite morphology bank inherit the bank's geometry as a structural prior. Richer banks --- expanded by collaborator-collected real measurements, augmented by physics-simulation, or replaced by a learned generative model --- would directly test whether more anchor diversity creates room for direction-of-perturbation to matter.

Second, the controlled-ablation methodology developed here is portable. The four-scenario design (baseline, random direction, label shuffle, inverted direction) requires no domain-specific knowledge once the bridge has explicit conditioning inputs and a perturbation mechanism. We recommend it as a default validation protocol for synthetic-data work that claims to encode any kind of label structure into outputs.

\subsection{Comparison to the QuaLiProM trajectory}

The QuaLiProM consortium~\cite{qualiprom} is building, in private, the kind of dataset MagBridge-Battery attempts to substitute for: paired magnetic measurements with electrochemical degradation labels at industrial scale. Our work and theirs occupy different positions on a spectrum: they pay the cost of measurement and gain ground-truth signal; we pay the cost of architectural assumptions and gain public availability. When QuaLiProM-style data eventually becomes public, MagBridge-Battery becomes immediately falsifiable in a constructive sense --- one can compare bridge outputs to real measurements at matched (SOH, SOC, voltage) tuples and quantify the mismatch.

\section{Limitations and intended use}
\label{sec:limitations}

\subsection{Intended uses}

MagBridge-Battery v1.0 is intended for:
\begin{itemize}
    \item Benchmark \textbf{SOH regression} on grounded-regime clean samples (T1).
    \item Benchmark \textbf{second-life classification} on grounded-regime clean samples (T2).
    \item Benchmark \textbf{anomaly detection} using the 600 synthetic anomalies and the 560 Regime-B extrapolation samples (T3).
    \item Studying \textbf{OOD extrapolation} via the three Regime-B voltage anchors.
    \item Quantifying \textbf{split-leakage effects} by training on both shipped splits and reporting the gap.
\end{itemize}

\subsection{Out-of-scope uses}

\begin{itemize}
    \item SOH regression on Regime-B samples (\texttt{soh} is \texttt{NaN} by design).
    \item Treating \texttt{B\_s1C5} or \texttt{B\_s2C6} as physical magnitudes.
    \item Treating \texttt{time\_norm} as per-sample timing information.
    \item Cross-chemistry transfer --- only LFP is represented in v1.0.
    \item Substituting MagBridge-Battery for real magnetic-sensing measurements in safety-critical decisions.
\end{itemize}

\subsection{Known limitations}

\textbf{Single-chemistry, single-cell coverage.} The v1.0 release uses LFP cells exclusively, and the OSF morphology bank is from one physical cell. The bridge therefore captures one cell's morphology, with no inter-cell variability in the magnetic signature.

\textbf{No ground-truth external validation.} The validation we report is structural (sanity invariants), distributional (KS tests at anchors), and ablation-based (A2). What we cannot do, by construction, is compare bridge outputs to real magnetic measurements from cells with known SOH labels --- because no such public data exists. The benchmark numbers reported here have meaning as internal metrics under documented bridge assumptions; their generalisation to real magnetic data will only be testable when paired data becomes available. Every generated sample carries its full provenance so that a future researcher with paired data can identify matching real samples and compute residuals directly.

\textbf{Regime-B carries no SOH label.} By design, the 560 Regime-B extrapolation samples have \texttt{soh}, \texttt{u\_features}, and \texttt{second\_life\_class} set to \texttt{NaN}. Users should filter to \texttt{regime == "grounded"} before fitting T1 / T2 models.

\textbf{Architectural ceiling from retrieve-and-blend decoder.} As discussed in \S\ref{sec:discussion:direction}, the A1$\approx$A0 finding partly reflects a property of the retrieve-and-blend decoder. A conditional generative-model decoder would in principle let SOH conditioning shape outputs non-trivially even with limited training data; this is the most promising single-axis future direction.

\textbf{Channel correlation structure.} The v1 bridge family treats per-channel perturbations as independent in time-domain noise application. At grounded anchors, the mean absolute difference between bridge and real channel-correlation matrices is approximately 0.4. This does not affect headline downstream-task results but is a known imperfection.

\textbf{Constant time axis.} \texttt{time\_norm} is the same vector for every sample (a fixed reference grid). It is included for loader convenience and carries no per-sample information.

\textbf{SOC coverage.} PulseBat samples cluster at SOC $\in \{5, 10, \ldots, 50\}\%$. Conditioning fidelity outside this range is untested.

\subsection{Future-work directions}

In rough order of cost-to-benefit:
\begin{enumerate}
    \item Conditional generative-model decoder (replace retrieve-and-blend with a small VAE or diffusion model on OSF samples).
    \item Lab collaboration for paired data --- even 50--100 cells with paired SOH labels and OPM measurements would enable ground-truth validation.
    \item Synthetic magnetic anchors via FEM simulation (expand the morphology bank).
    \item Chemistry-transfer extension once paired data exists at any chemistry beyond LFP.
    \item Integration with QuaLiProM-style data when it becomes public.
\end{enumerate}

\section{Conclusion}
\label{sec:conclusion}

We released \textbf{MagBridge-Battery v1.0}, a 6{,}760-sample synthetic dataset bridging the OSF magnetic-morphology archive with PulseBat electrochemical degradation labels, motivated by the absence of any public dataset pairing magnetic measurements with state-of-health information. The release comprises 5{,}600 PulseBat-conditioned grounded samples, 600 synthetic sensor-anomaly samples derived from clean parents, and 560 low-voltage Regime-B extrapolation samples, together with a cell-disjoint, parent-child-leakage-free primary benchmark split. A label-shuffle controlled ablation confirms that the bridge encodes input SOH into output morphology non-trivially: SOH regression collapses from $R^2 \approx 0.77$ to $\approx 0$ when the SOH--cell binding is broken at generation time. A direction-perturbation ablation establishes a complementary architectural finding: in a retrieve-and-blend bridge over a small, sharply-clustered morphology bank, the principled latent direction is not privileged for downstream decoding, because the bank's geometry dominates the signal. The dataset, bridge implementation, and a benchmark suite covering SOH regression, second-life classification, and anomaly detection are released to support method development while public paired magnetic--electrochemical data remains unavailable.

\section*{Code and data availability}

MagBridge-Battery v1.0 is released on Zenodo under CC-BY-4.0 (DOI: \texttt{10.5281/zenodo.20260147}). The bridge implementation, ablation scripts, and benchmark suite are released on GitHub under Apache-2.0 at \url{https://github.com/SakthiGs/MagBridge-Battery}. The release bundle contains data shards (Parquet), metadata, both benchmark splits, the generation manifest pinning bridge version and configuration hash, a minimal Python loader, SHA-256 checksums, a citation file, dataset documentation, and a LICENSE file documenting the upstream sources (OSF magnetometry archive, PulseBat dataset) and their respective license declarations.

\section*{How to cite}

Users of MagBridge-Battery are kindly requested to cite both the dataset DOI and this paper. The dataset DOI uniquely identifies the v1.0 data artifact; the paper documents the bridge construction, validation, and benchmark protocol. See \texttt{CITING.md} in the release bundle for copy-paste BibTeX.

\section*{Acknowledgements}

The authors thank the Mohammadi--Jerschow group for openly releasing the OSF battery magnetometry archive, which made this work possible. We also thank Prof. Alexej Jerschow for helpful clarification regarding the public OSF dataset. The authors further thank the PulseBat team (Tao et al.) for releasing pulse-voltage response data on retired Li-ion cells. Any errors or interpretations in this work remain the responsibility of the authors.

\appendix
\section{Bridge degradation modulator: full equations}
\label{app:archmath}

For completeness we list the equations referenced in \S\ref{sec:arch:modulator}. The morphology bank produces a base sample at voltage $v$ by linear interpolation between bracketing anchors $v_l \leq v \leq v_h$:
\begin{equation}
    \boldsymbol{\mu}(v) = (1-\alpha)\,\boldsymbol{\mu}_{v_l} + \alpha\,\boldsymbol{\mu}_{v_h}, \quad \alpha = \frac{v - v_l}{v_h - v_l}
\end{equation}
\begin{equation}
    \boldsymbol{\sigma}(v) = \sqrt{(1-\alpha)\,\boldsymbol{\sigma}_{v_l}^2 + \alpha\,\boldsymbol{\sigma}_{v_h}^2}
\end{equation}
\begin{equation}
    \mathbf{X}_{\text{base}} = \boldsymbol{\mu}(v) + \boldsymbol{\sigma}(v) \odot \boldsymbol{\epsilon}, \quad \boldsymbol{\epsilon} \sim \mathcal{N}(0, I)
\end{equation}
The degradation modulator computes the MagBridge-Embed of the base sample, projects to LDA space:
\begin{equation}
    \mathbf{z}_{\text{base}} = (\boldsymbol{\Sigma}^{-1/2}(\mathbf{e}_{\text{base}} - \boldsymbol{\mu}_e) - \bar{\mathbf{x}}) \mathbf{W}_{\text{LDA}}
\end{equation}
perturbs along the fitted state direction $\hat{\mathbf{d}}$:
\begin{equation}
    \mathbf{z}_{\text{pert}} = \mathbf{z}_{\text{base}} + m \cdot \hat{\mathbf{d}}, \quad m = -\gamma \cdot \delta, \quad \delta = \max(0, 1 - \text{SOH})
\end{equation}
decodes via cone-restricted $k$-NN softmin:
\begin{equation}
    \mathbf{X}_{\text{decode}} = \sum_{i \in \mathcal{N}_k} w_i \cdot \mathbf{X}_i^{\text{OSF}}, \quad w_i = \frac{\exp(-d_i / \tau)}{\sum_j \exp(-d_j / \tau)}
\end{equation}
and blends with the base proportionally to $\delta$:
\begin{equation}
    \mathbf{X}_{\text{mod}} = (1 - \beta)\,\mathbf{X}_{\text{base}} + \beta\,\mathbf{X}_{\text{decode}}, \quad \beta = \min(1, 1.5\delta)
\end{equation}
with v1.0 release configuration $\gamma = 800$, $k=8$, $\tau = 50$, cone half-angle $75^\circ$. The bridge configuration is pinned by hash in the released manifest.

\bibliographystyle{IEEEtran}
\bibliography{references}

\end{document}